%% file: main.tex
\documentclass[10pt,twocolumn,letterpaper]{article}

\usepackage{iccv}              

\input{preamble}

\definecolor{iccvblue}{rgb}{0.21,0.49,0.74}
\usepackage[pagebackref,breaklinks,colorlinks,allcolors=iccvblue]{hyperref}

\def\paperID{8880} 
\def\confName{ICCV}
\def\confYear{2025}

\title{FedORGP: Guiding Heterogeneous Federated Learning with Orthogonality Regularization on Global Prototypes}

\author{
  Fucheng Guo\thanks{Shenzhen International Graduate School, Tsinghua University, \tt\small gfc23@mails.tsinghua.edu.cn}, 
  Zeyu Luan\thanks{Peng Cheng Laboratory, China}, 
  Qing Li\thanks{Peng Cheng Laboratory, China, \tt\small liq@pcl.ac.cn}, 
  Dan Zhao\thanks{Peng Cheng Laboratory, China, \tt\small zhaod01@pcl.ac.cn}, 
  Yong Jiang\thanks{Shenzhen International Graduate School, Tsinghua University, \tt\small liq@pcl.ac.cn}
}

\begin{document}

\maketitle
\input{sec/abstract}
\input{sec/introduction}
\input{sec/related_work}
\input{sec/methodology}
\input{sec/Convergence_Analysis}
\input{sec/Experiments}
\input{sec/conclusion}
{
    \small
    \bibliographystyle{ieeenat_fullname}
    \bibliography{main}
}

\end{document}

%% file: preamble.tex
\usepackage{algorithm}
\usepackage{algpseudocode}
\usepackage{amsmath}
\usepackage{amssymb}
\usepackage{pifont}

%% file: sec/abstract.tex
\begin{abstract}
Federated Learning (FL) has emerged as an essential framework for distributed machine learning, especially with its potential for privacy-preserving data processing. However, existing FL frameworks struggle to address statistical and model heterogeneity, which severely impacts model performance. While Heterogeneous Federated Learning (HtFL) introduces prototype-based strategies to address the challenges, current approaches face limitations in achieving optimal separation of prototypes. This paper presents FedORGP, a novel HtFL algorithm designed to improve global prototype separation through orthogonality regularization, which not only encourages intra-class prototype similarity but also significantly expands the inter-class angular separation. With the guidance of the global prototype, each client keeps its embeddings aligned with the corresponding prototype in the feature space, promoting directional independence that integrates seamlessly with the cross-entropy (CE) loss. We provide theoretical proof of FedORGP's convergence under non-convex conditions. Extensive experiments demonstrate that FedORGP outperforms seven state-of-the-art baselines, achieving up to 10.12\% accuracy improvement in scenarios where statistical and model heterogeneity coexist.
\end{abstract}

%% file: sec/introduction.tex
\section{Introduction}
\label{sec:introduction}

In recent years, Federated Learning (FL) has emerged as a promising distributed machine learning paradigm \cite{promising}. FL eliminates the need for clients to exchange data, allowing data to remain decentralized, which makes it an favorable solution to data privacy challenges \cite{uncentralized, fedavg}.

Previous work mainly focuses on homogeneous FL, which assumes that all client models are identical. However, in large-scale real-world scenarios, considerable disparities exist in both client data distribution, often termed statistical heterogeneity, and hardware resources, known as system heterogeneity \cite{data_and_model_heterogenous, FedProx}. Data heterogeneity severely impacts the convergence and performance of global model \cite{model-training-on-noniid-data, fl_challenges_methods_directions}. Furthermore, discrepancies in hardware resources can result in model heterogeneity across clients, posing significant challenges to conventional FL approaches that rely on aggregating model parameters \cite{data_and_model_heterogenous, FLadvance}. In addition, homogeneous FL trains a shared global model by exchanging gradients, which further imposes significant communication costs as well as privacy exposure risks \cite{aggregating-vs-privacy, fedtgp}.

\begin{figure*}[t]
    \centering
    \begin{subfigure}[b]{0.33\linewidth}
        \includegraphics[width=\linewidth]{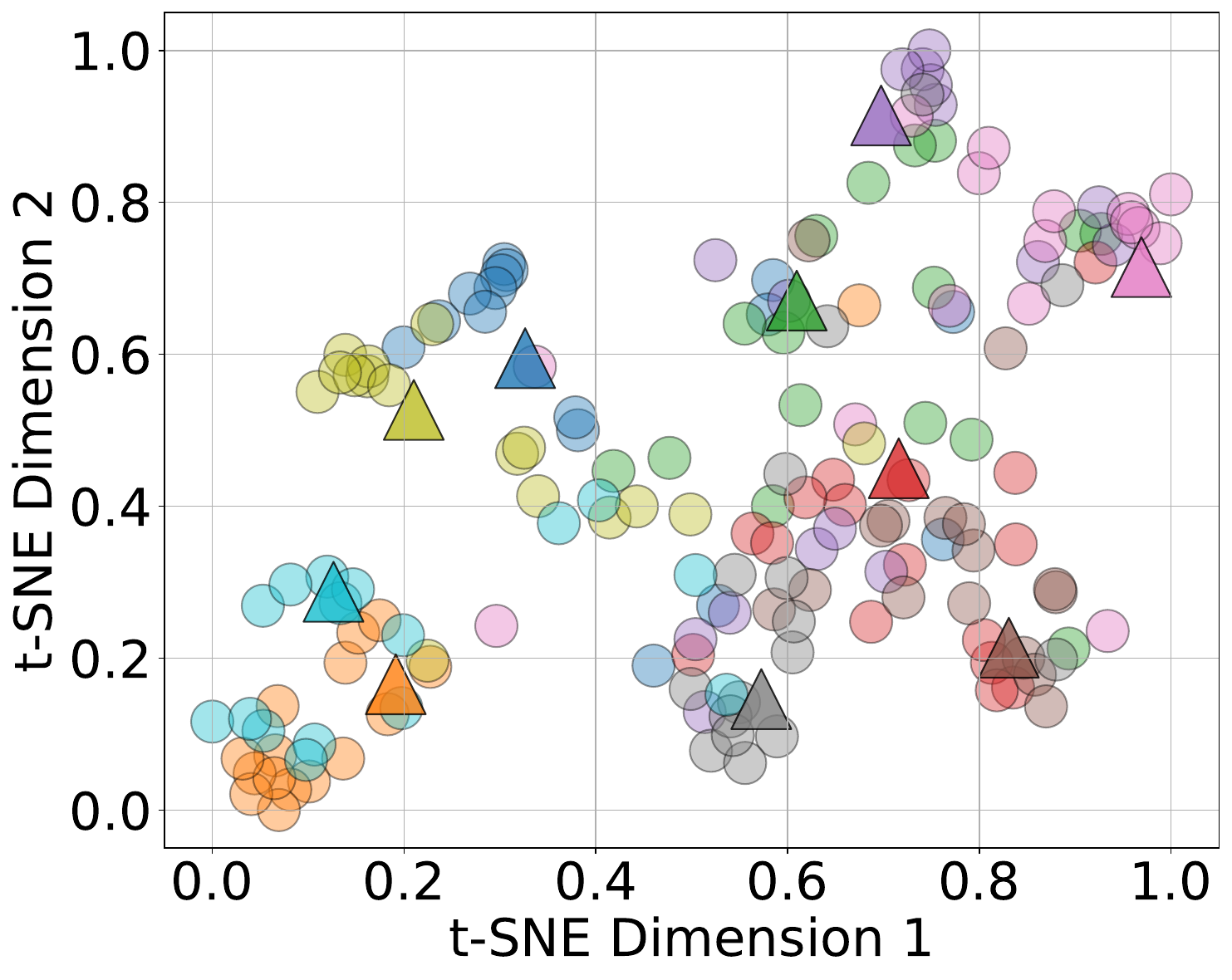}
        \caption{FedProto}
        \label{fig:FedProto t-SNE}
    \end{subfigure}
    \begin{subfigure}[b]{0.33\linewidth}
        \includegraphics[width=\linewidth]{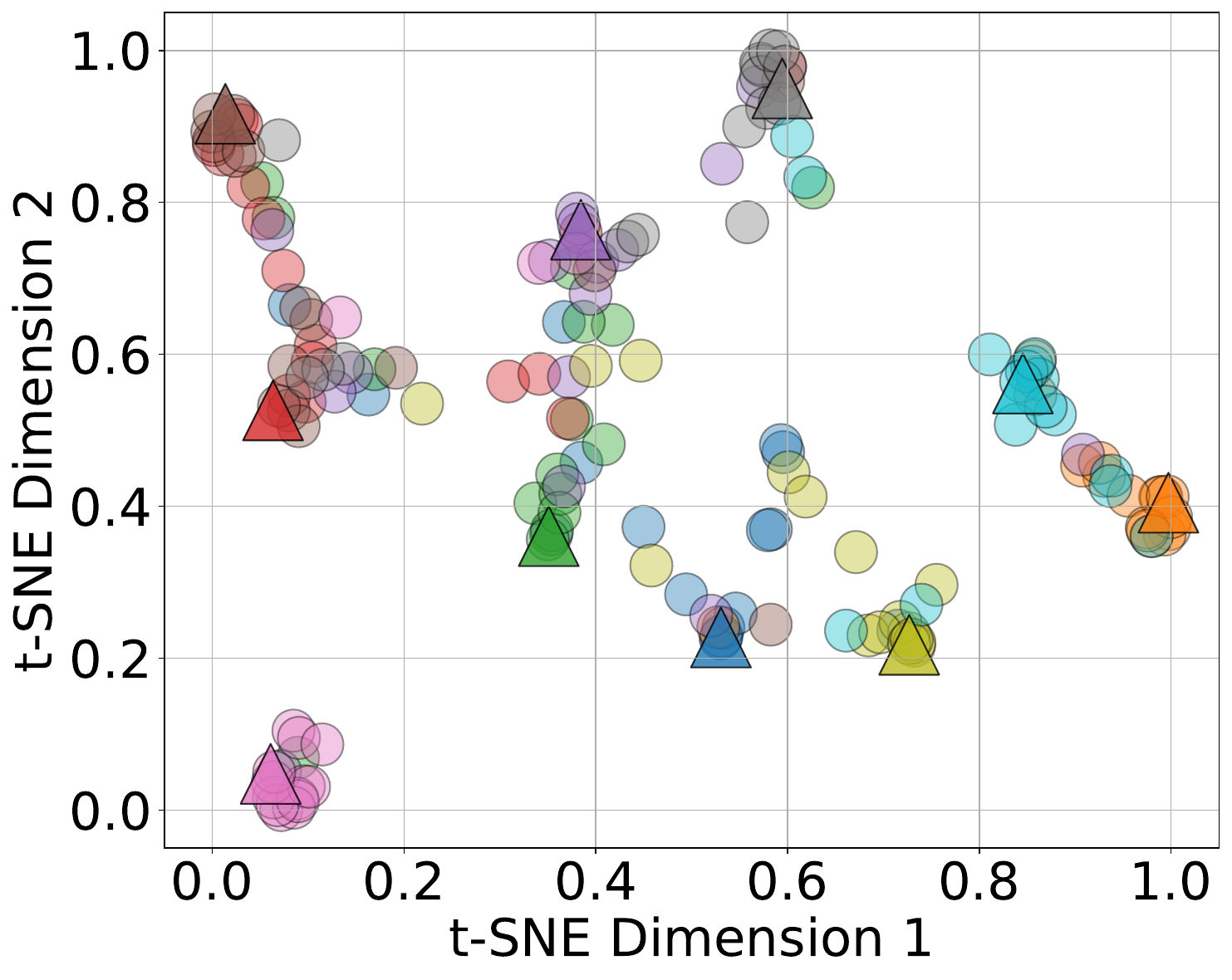}
        \caption{FedTGP}
        \label{fig:FedTGP t-SNE}
    \end{subfigure}
    \begin{subfigure}[b]{0.33\linewidth}
        \includegraphics[width=\linewidth]{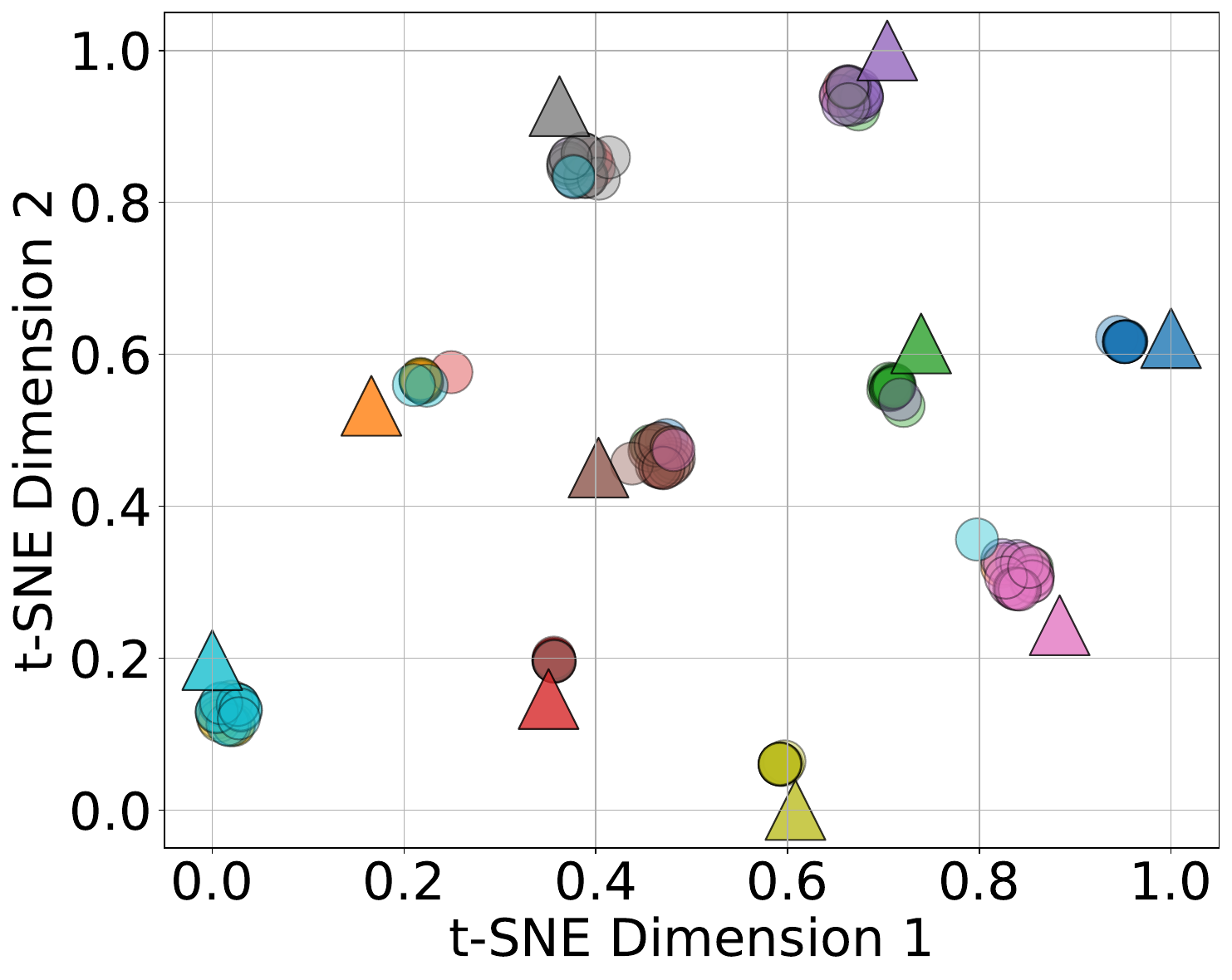}
        \caption{FedORGP}
        \label{fig:FedORGP t-SNE}
    \end{subfigure}
    \caption{We train models on the CIFAR-10 dataset and use t-SNE \cite{t-sne} to visualize their performance on previously unseen test samples (16 per class) within the feature space, with triangles indicating prototypes and circles denoting samples. The results indicate that FedProto \cite{tan2022fedproto} exhibits weak feature separation. FedTGP \cite{fedtgp} increases prototype margin but still lacks sufficient distinction in the feature space. In contrast, our FedORGP can reduce inter-sample similarity and increase intra-class compactness, thereby effectively classifying the majority of samples. This result suggests that FedORGP exhibits robust generalization performance under conditions of both statistical and model heterogeneity.}
    \label{fig:distributions}
\end{figure*}

To tackle these challenges, Heterogeneous Federated Learning (HtFL) has emerged as a novel FL paradigm capable of handling both data heterogeneity and model heterogeneity simultaneously \cite{tan2022fedproto, yi2023fedgh, fedtgp}. HtFL incorporates prototype-based learning, which communicates client prototypes rather than model gradients to the server, thus alleviating issues related to model diversity and communication costs. However, existing weighted average HtFL solutions like FedProto \cite{tan2022fedproto} faces several limitations. First, aggregating global prototypes on the server side via weighted averaging requires clients to upload sample sizes, which may lead to leakage of data distribution information \cite{yi2023fedgh}. Second, as shown in Fig.\ref{fig:FedProto t-SNE}, the weighted average prototype lacks a well-defined decision boundary, resulting in overlapping feature distributions among different classes in the feature space \cite{fedtgp}.

While the recent HtFL solution FedTGP \cite{fedtgp} has successfully improved separation by increasing the Euclidean distance between prototypes, this approach still faces several limitations. First, contrastive learning, which works together with cross-entropy (CE) loss, primarily reduces the intra-class distance between prototypes but fails to explicitly enforce inter-class separation. Second, as shown in Fig.\ref{fig:FedTGP t-SNE}, in high-dimensional spaces, the Euclidean distance makes it difficult to effectively distinguish different samples as distance variations between samples diminish \cite{comprehensive-survey-of-distance}. Third, augmenting Euclidean distances fails to effectively leverage the angular separation characteristics inherent in CE loss \cite{opl}.

To address these limitations, we propose a novel HtFL algorithm called FedORGP, which optimizes global prototypes through orthogonality regularization. FedORGP improves intra-class prototype similarity while preserving semantic integrity and explicitly promotes angular separation between classes, ensuring that global prototypes achieve maximal directional independence in the feature space. Under the guidance of global prototypes, as demonstrated in Fig.\ref{fig:FedORGP t-SNE}, clients can ensure that the feature representations are directionally aligned with their corresponding global prototypes in the feature space, thereby facilitating a more effective integration with the angular properties inherent to CE loss.

We evaluate FedORGP against seven state-of-the-art HtFL methods across four datasets in scenarios where both data heterogeneity (practical distribution and Dirichlet distribution) and model heterogeneity (four levels of model heterogeneity) coexist. The experimental results show an improvement in accuracy by 10.12\% over the best baseline under both statistical and model heterogeneity. Our contributions can be summarized as follows:

\begin{itemize}
    \item We present a novel framework FedORGP for HtFL with orthogonality regularization. To the best of my knowledge, this is the first work to introduce orthogonality regularization into heterogeneous federated learning where data and model heterogeneity coexist.
    \item We observe that increasing the Euclidean distance between prototypes is incapable of effectively integrating with the angular characteristics inherent to CE loss. Our proposed FedORGP algorithm enhances intra-class clustering within the feature space while explicitly promoting angular separation between classes.
    \item We offer a theoretical convergence guarantee for FedORGP and rigorously establish the convergence rate under non-convex conditions.
\end{itemize}

%% file: sec/related_work.tex
\section{Related Work}
\label{sec:related work}

\subsection{Heterogenous Federated Learning}

Heterogeneous Federated Learning (HtFL) has emerged as a solution that supports both model and statistical heterogeneity simultaneously, while ensuring user privacy.
Early HtFL solutions \cite{diao2020heterofl, horvath2021fjord, wen2022federated} let clients sample different submodels from a shared global architecture, thereby reducing the computational burden on low-resource devices. However, these methods necessitate a common architecture to be shared across clients, raising privacy concerns regarding clients’ model architectures.

An alternative approach to system heterogeneity shares a common model of the same structure at the server side, e.g. a global header \cite{yi2023fedgh} or a global generator \cite{FedGen}, as a way to transfer global knowledge. LG-FedAvg \cite{lgfedavg} and FedGH \cite{yi2023fedgh} require clients to share the same top layer while allowing them to have different architectures in their lower layers. However, sharing and aggregation of top layers may result in suboptimal performance due to statistical heterogeneity \cite{nofearofclassifierbiases, nofearofheterogenity}. FedGen \cite{FedGen} is designed to mitigate model drift in heterogeneous FL environments by using a generative model to extract global knowledge from local models. While this method avoids reliance on proxy datasets, its performance highly relies on the quality of the generator \cite{fedtgp}. Despite allowing some level of flexibility, they still assume that clients share partially similar models. 

Alternatively, another HtFL approach aims to enable completely independent client model architectures by exchanging various forms of information, such as knowledge or prototypes. FedMD \cite{li2019fedmd} and FedDF \cite{FedDF} facilitate knowledge transfer between participants through distillation over public datasets. However, ideal public datasets are often difficult to access \cite{public_dataset_is_difficult_to_obtain}. FML \cite{fml} and FedKD \cite{fedkd} train and share a small auxiliary model using mutual distillation, circumventing the need for a global dataset \cite{fedtgp}. Nonetheless, the bi-directional distillation process relies on frequent communication between clients and the server, which can result in significant communication overhead.

Prototype-based HtFL methods \cite{tan2022fedproto,FedPCL,FedPAC} have shown promise in addressing both model and statistical heterogeneity. These approaches not only enhance model convergence in non-IID settings but also substantially reduce communication overhead by transmitting prototypes instead of full models \cite{FedPCL}. However, these methods rely on simple weighted averages of client prototypes, which increase the chance of overlap between global prototypes in the feature space. While FedTGP \cite{fedtgp} increases prototype separation by contrastive learning without using weighted averages, boosting the Euclidean distance between embeddings does not integrate well with CE loss \cite{opl, comprehensive-survey-of-distance}. In this work, we introduce orthogonal regularization to explicitly enhance the angular margin among global prototypes, thereby tackling data and model heterogeneity in HtFL.

\subsection{Prototype Learning}
Prototype refers to the average of features representations of the same class. FedNH \cite{FedNH} addresses data heterogeneity by leveraging uniformity and semantics of class prototypes. FPL \cite{FPL} and FedPLVM \cite{FedPLVM} mainly focus on solving the domain drift problem using prototype clustering. However, they all require model aggregation, which necessitates that client models share the same architecture, implying that they cannot work in scenarios with model heterogeneity. Our proposed FedORGP encourages prototype separation to establish distinct margins among prototypes, thereby guiding heterogeneous client models to assimilate global knowledge.

\subsection{Orthogonality}
Orthogonality typically describes the directional independence between two vectors. In high-dimensional space, two vectors are considered orthogonal if their dot product is zero \cite{opl}. FediOS \cite{Fedios} effectively separates generic and personalized features using orthogonal projections to address feature skew in personalized federated learning. However, it requires all client models to have identical architectures. C-FSCIL \cite{C-FSCIL}, OPL \cite{opl} and POP \cite{POP} enhance feature separation between classes via orthogonality regularization. However, they are only applicable to the optimisation of centralised training scenarios and are not applicable to the optimisation of federated scenarios. FOT \cite{FOT} and FedSOL \cite{FedSOL} are primarily concerned with the orthogonality of update directions to minimise interference across clients or tasks. However, neither of them can guarantee the orthogonality of the samples in the feature space.

%% file: sec/methodology.tex
\section{Methodology}
\label{sec:methodology}

\subsection{Problem Definition}
We consider a problem involving $M$ clients, which have heterogeneous models and private data. The local model of client $k$, $k\in\{1,\dots,M\}$, is split into two components: a feature extractor $f_k$ parameterized by $\phi_k$, and a classifier $h_k$ parameterized by $\theta_k$. Given a sample pair ($x, y$), the feature extractor $f_k$ transforms $x \in \mathbb{R}^{D}$ into a feature representation $r=f_k(x;\phi_k)$, $r \in \mathbb{R}^{K}$, where $D$ is the dimension of input and $K$ is the dimension of feature representation ($K \ll D$). The classifier $h_k$ maps the feature vector $r$ to logits $\in \mathbb{R}^{C}$, where $logits = h_k(r; \theta_k)$ and $C$ denotes the total number of classes. The client model's parameters are defined as $\omega_k = (\phi_k, \theta_k)$. The optimization objective of FedORGP is defined as:
\begin{equation}\min\sum_{k=1}^M\frac{|\mathcal{D}_k|}{|\mathcal{D}|}\mathcal{L}_k\left(\mathcal{D}_k,\omega_k,\mathcal{P}\right),
    \label{eq:overall objective}
\end{equation}
where $|\mathcal{D}|$ denotes the total size of datasets across all clients, $|\mathcal{D}_k|$ indicates the size of the dataset of client $k$, $\mathcal{P}$ represents the global prototype.

\subsection{Orthogonality Regularization on Prototypes}
Following FedProto \cite{tan2022fedproto}, the prototype of class $c$, $c \in \{1, \ldots, C\}$, for client $k$ is defined as follows:
\begin{equation}
    P_k^c=\mathbb{E}_{(x,c)\sim\mathcal{D}_{k,c}}f_k(x;\phi_k),
    \label{eq:client prototype definition}
\end{equation}
where $D_{k, c}$ is the subset of $D_k$ consisting of samples of class $c$. We initialize a random embedding $\tilde{P}^c$ for each class $c$. We define the trainable prototypes module $F$ which comprises two Fully-Connected layers with a ReLU activation function in between, parameterized by $\omega_s$ \cite{fedtgp}. We input the initial prototype vector $\tilde{P}^c\in\mathbb{R}^K$ into $F$, producing the final global prototype $\hat{P}^c=F(\tilde{P}^c;\omega_s)$. The transformation network parameters, along with the embeddings, are optimized jointly during training. This module is designed to generate adaptable prototype for each class.

To enhance the separation between global prototypes in the feature space, we adopt orthogonality regularization to train global prototypes. After client local training, the server will get the client prototypes $\mathcal{P}_k$, which consists of the prototypes of different classes of $k$-th client. Then, the dataset we use to train the global prototype can be expressed as $\mathcal{Q} = \bigcup_{k=1}^{M} \{ P^c_k \mid c \in \mathcal{C}_k \}$, where $\mathcal{C}_k \subseteq \{0, 1, ..., C - 1\}$ denotes the set of classes present on $k$-th client.

Our purpose is to cluster $\hat{P}^c$ and $P_k^c$, while discriminating between $\hat{P}^c$ and $P_k^{\Bar{c}}$, where $\bar{c} \ne c$. Formally, within a mini-batch $B_p \subseteq \mathcal{Q}$, we define $s$ as intra-prototype similarity and $d$ as inter-prototype similarity:
\begin{align}
    s &= \frac{1}{|B_p|}\sum_{P_k^c \in B_p}\langle P_k^c, \hat{P}^c\rangle, 
    \label{eq:similarity eqution}\\[1ex]
    d &= \frac{1}{|B_p|}\frac{1}{|C| - 1}\sum_{P_k^c \in B_p}\sum_{\substack{\bar{c} \in C \\ c\ne\bar{c}}}\left|\langle P_k^c, \hat{P}^{\bar{c}}\rangle\right|,
    \label{eq:dissimilarity eqution}
\end{align}
where $k \in \{1, \ldots, m\}$, $m$ is the number of clients participating in the training, $\left|\cdot\right|$ is the absolute value operator. Note that the cosine similarity operator $\langle\cdot,\cdot\rangle $ on two vectors involves normalization of features (projection to a unit hypersphere) and is calculated as $\langle v_i, v_j \rangle=\frac{v_i\cdot v_j}{\|v_i\|_2\cdot\|v_j\|_2}$, where $||\cdot||_2$ refers to the $l_2$ norm operator.

Then, we define a unified loss function $\mathcal{L}_{\mathrm{OR}}$ that simultaneously ensures intra-class clustering and inter-class orthogonality within a mini-batch as follows:
\begin{equation}
    \mathcal{L}_{\mathrm{OR}}=\lambda_s*(1-s)+\gamma*d,
    \label{eq:OC Loss eqution}
\end{equation}
where $\lambda_s$ and $\gamma$ are hyperparameters to balance the contributions of each term to the overall loss. Specifically, $\lambda_s$ is employed to regulate the influence of the intra-class loss, while $\gamma$ controls the weight of the inter-class loss. This configuration allows for a flexible adjustment of the intra-class compactness and inter-class separation.

\begin{figure}[t]
    \centering
    \includegraphics[width=\linewidth, height=3.3cm]{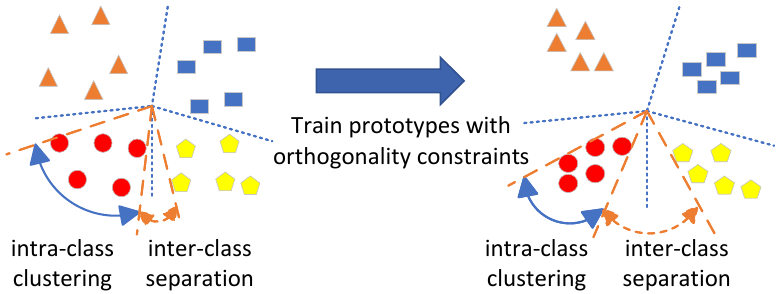}
    \caption{Orthogonality-constrained prototypes can expand inter-class margins while enhancing intra-class clustering.}
    \label{fig:orthogonal constraint picture}
\end{figure}

The training objective is to minimize the loss function $\mathcal{L}_{\mathrm{OR}}$ defined in Eq. \ref{eq:OC Loss eqution}. Notice that $(1 - s) > 0$ inherently holds (as $s \in (-1, 1)$, making $(1 - s) \in (0, 2)$), and $d$ is already an absolute value with $d \in (0, 1)$. Therefore, we should maximize $s$ toward 1 while minimizing $d$ toward 0. 
As depicted in Fig.\ref{fig:orthogonal constraint picture}, when minimizing this overall loss, the first term $\lambda_s * (1 - s)$ promotes clustering of prototypes within the same class, while the second term $\gamma * d$ enforces orthogonality among prototypes of different classes. The loss can be implemented efficiently in a vectorized manner on the mini-batch level, avoiding any loops.

To maintain directional independence, the prototypes of different classes are constrained to be orthogonal during the optimization process. This structure not only improve intra-class compactness but also explicitly enlarges inter-prototype separation, while preserving the semantic integrity of each class. 

\subsection{The Suitability of Orthogonality for CE Loss}
The multi-class Cross-Entropy loss function is a common loss function used in classification problems, for each sample $(x, y)$, the loss function can be defined as follows: 
\begin{align}
    \mathcal{L}_{CE}(\hat{y}, y) 
    &= -\sum_{c=1}^{C} \mathbb{I}(c = y) \log(p_c)
    \nonumber\\
    &= - \log\left(\frac{\exp(z_{y})}{\sum_{c' = 1}^{C} \exp(z_{c'})} \right)\\
    &= - \log \left( \frac{\exp(r^y \cdot w^y)}{\sum_{c' = 1}^{C} \exp(r^y \cdot w^{c'})} \nonumber\right)
     \label{eq:cross_entropy}
\end{align}
where $\hat{y}$ is the predicted label and $y$ is the truth label. $\mathbb{I}\left(\cdot\right)$ is an indicator function. $p_{c}$ represents the predicted probability that the model predicts the sample to belong to class $c$. $z_y$ is the logits of the sample. 
We define the classifier $W = [w^1, ... , w^C]$, where $w^y \in \mathbb{R}^K$ is the learning projection vector of label $y$. The CE loss can then be defined in terms of discrepancy between the predicted $\hat{y}$ and ground-truth label $y$, by projecting the features $r^y$ onto the weight matrix $W$. When a client trains the model with the global prototypes, it ensures that $r^y$ remains aligned with the prototype $\hat{P}^y$. During updates with SGD, the CE loss further reinforces the alignment between weight vector $w^y$ and its corresponding feature representation $r^y$ through a dot product operation, given as: 
\begin{equation}
r^y \cdot w^y=\|r^y\|\cdot\|w^y\|\cdot\cos(\theta_{r^y,w^y}),
\label{eq:dot product}
\end{equation}
where $\theta_{r^y,w^y}$ represents the angle between $r^y$ and $w^y$. Thus, when the angle between the feature representation $r^y$ and the weight vector $w^y$ approaches zero, their inner product is maximized, indicating directional alignment. Meanwhile, since the feature representation $r^y$ is explicitly aligned with the global prototype $\hat{P}^y$ during training, the three components ($r^y$, $w^y$, and $\hat{P}^y$) continuously align with each other in the feature space. However, the Cross-Entropy (CE) loss inherently lacks an explicit mechanism to enforce angular separation between different classes, leading to potential overlap in feature representations and consequently suboptimal discriminative performance. To overcome this limitation, we propose orthogonality regularization (OR) to explicitly enlarge angular separation among global prototypes. By enforcing orthogonality among global prototypes, OR significantly reduces inter-class overlap, thereby ensuring robust directional consistency among $r^y$, $w^y$, and $\hat{P}^y$, and ultimately improving the discriminative capability of the classifier in heterogeneous federated environments.

\begin{figure}[t]
    \centering
    \includegraphics[width=\linewidth]{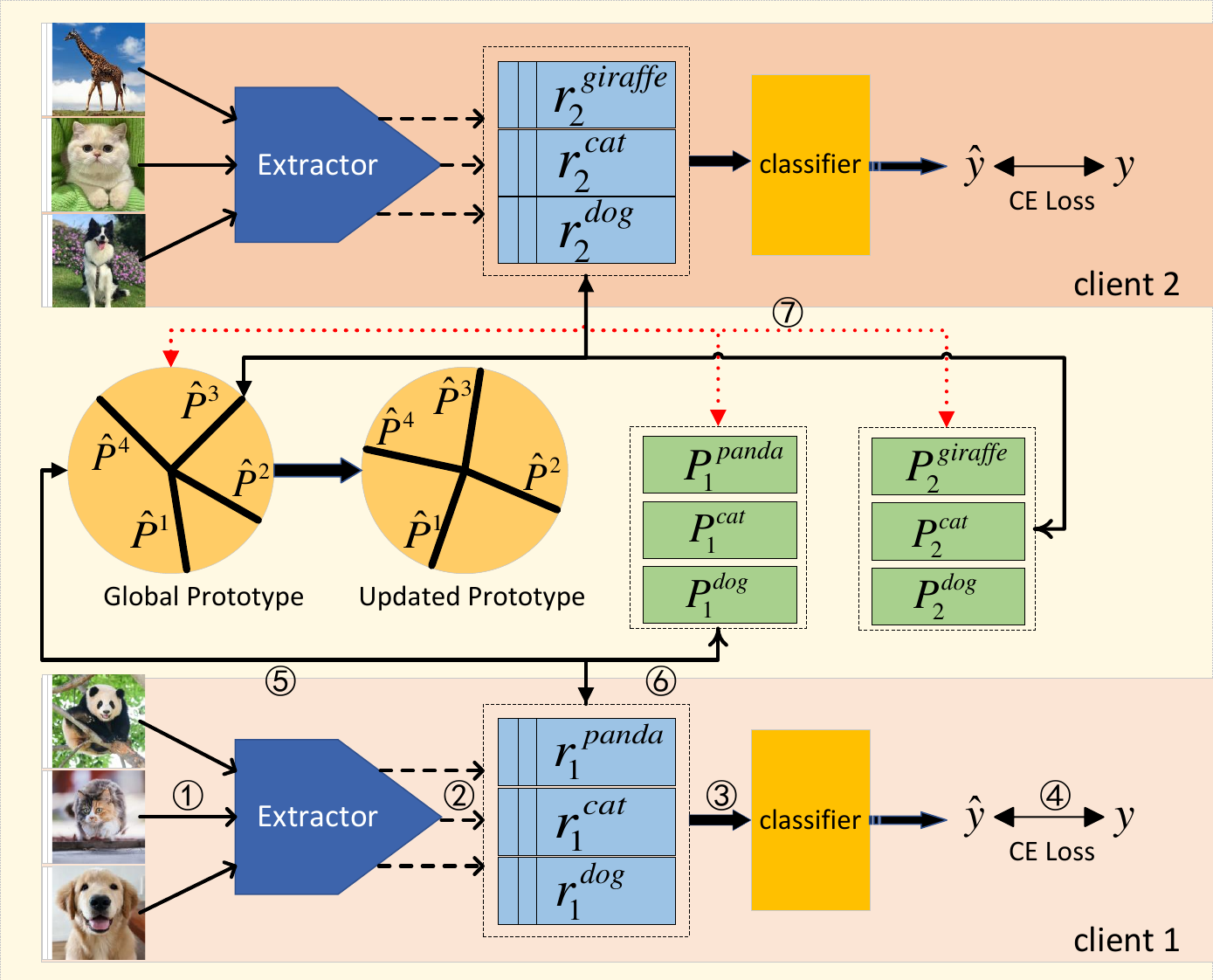}
    \caption{This image illustrates the FedORGP framework involving two clients. From \ding{172} to \ding{175}: We input the samples, get the representations, then input them into the classifier for prediction, and finally calculate the CE loss. \ding{176}: The dissimilarity between the feature representation and its corresponding prototype is calculated as a regularization term. \ding{177}: After the local model is updated, we collect the client prototypes. \ding{178}: Orthogonality regularization are applied to the global prototypes on the server to enhance angular separation.}
    \label{fig:overview}
\end{figure}


\subsection{Local Model Update}
In FedORGP, the client updates its local model to generate embeddings that are consistent with the global prototypes shared across clients. To maintain directional alignment with the global prototype, a regularization term is added to the local loss function, which encourages feature representations from $k$-th client $r_k^c$ to align with their respective global prototypes $\hat{P}^c$ while minimizing classification error. Then, the client prototype $P_k^c$ has a more separate margin. Specifically, the loss function is formulated as follows:
\begin{equation}
\begin{split}
    \mathcal{L}_k:=
    &\mathbb{E}_{(x,y)\sim \mathcal{D}_k}\mathcal{L}_{CE}(h_k(f_k(x;\phi_k);\theta_k),y)+\\
    &\lambda_c \cdot \mathbb{E}_{(x,y)\sim \mathcal{D}_k}\mathcal{L}_R(f_k(x;\phi_k),\hat{P}^y),
\end{split}
\label{eq:client loss eqution}
\end{equation}
where $\lambda_c$ is a hyperparameter.
This regularization $\mathcal{L}_R$ facilitates the alignment between feature representations and global prototypes across clients, enhancing the global consistency and robustness of the federated model in heterogeneous environments.

\subsection{FedORGP Framework}

\begin{algorithm}[t]
\caption{The learning process of FedORGP}
\label{alg:workflow}
\begin{algorithmic}[1]
    \Require $M$ clients with heterogeneous models and data, trainable global prototypes $\hat{\mathcal{P}}$ on the server, local epochs $E$, and total communication rounds $T$
    \Ensure Well-trained client models.

    \State \textbf{Server:}
    \State \textbf{for} round $t = 1, \dots, T$ \textbf{do}
        \State \quad Randomly sample a client subset $\mathcal{I}^t$
        \State \quad Send $\hat{\mathcal{P}}$ to clients in $\mathcal{I}^t$
        \State \quad \textbf{for each} Client $k \in \mathcal{I}^t$ \textbf{in parallel do}
            \State \quad\quad Client $k$ performs local training
        \State \quad \textbf{end for}
        \State \quad Train the global prototype on the server by Eq. \ref{eq:OC Loss eqution}
    \State \textbf{end for}
    \State \Return Well-trained client models.

    \State \textbf{Client Local Training:}
    \State \quad \textbf{for} iteration $e$ = 1, \ldots, $E$ \textbf{do}
    \State \quad \quad Local training with the global prototype by Eq. \ref{eq:client loss eqution}
    \State \quad \textbf{end for}
    \State \quad Collect local prototypes $\{P^c_k | c \in \mathcal{C}_k\}$ by Eq. \ref{eq:client prototype definition}
    \State \quad Send local prototypes $\mathcal{P}_k$ to the server
\end{algorithmic}
\end{algorithm}

We present the complete workflow of FedORGP in Alg.\ref{alg:workflow} and give an illustration in Fig.\ref{fig:overview}. The well-trained, class-separable global prototypes are then distributed to clients in the subsequent round to guide client training for improving separability among feature representations.

%% file: sec/Convergence_Analysis.tex
\section{Convergence Analysis}
\label{sec:convergence analysis}

To analyse the convergence of FedORGP, we first introduce some additional notes as in the existing framework. Unless otherwise stated, we always write the client-side loss function $\mathcal{L}_k(\mathcal{D}_k, \omega_k, \mathcal{P})$ as $\mathcal{L}_k$. In order to analyse the convergence of FedORGP, we first introduce some additional assumptions \cite{tan2022fedproto, yi2023fedgh, FedNH}.\\
\textbf{Assumption 1.} (Lipschitz Smooth). The k-th client's gradient of local loss function is $L_1$-Lipschitz continuous:
\begin{equation}
    \|\nabla\mathcal{L}_k^{t_1}-\nabla\mathcal{L}_k^{t_2}\|_2 
    \leqslant L_1\|\omega_k^{t_1}-\omega_k^{t_2}\|_2, 
    \label{eq:lipschitz-continuous}
\end{equation}
where $\forall t_1, t_2 > 0, k \in \{1, \dots, m\}$. That also means that local objective function is $L_1$-Lipschitz smooth:
\begin{equation}
        \mathcal{L}_k^{t_1} - \mathcal{L}_k^{t_2} \leqslant \langle \nabla \mathcal{L}_k^{t_2}, (\omega_k^{t_1} - \omega_k^{t_2}) \rangle + \frac{L_1}{2} \|\omega_k^{t_1} - \omega_k^{t_2}\|_2^2,
    \label{eq:lipschitz-smooth}
\end{equation}
\textbf{Assumption 2.} (Unbiased Gradient and Bounded Variance). The stochastic gradient $g_{k}^t=\nabla\mathcal{L}(\xi_i, \omega^t)$ is an unbiased estimator of the local gradient for k-th client, where $\xi_i$ is a random variable representing the randomness in the data (e.g., mini-batch). Suppose its expectation is
\begin{equation}
    \mathbb{E}_{\xi_i\thicksim D_i}[g_{k}^t]=\nabla\mathcal{L}_k(\omega_{k}^t),\forall k\in\{1,2,\ldots,m\},
    \label{eq:unbiased gradient}
\end{equation}
and there exists $\sigma^2 \geq 0$, then the variance of random gradient $g_{k}^t$ is bounded by:
\begin{equation}
        \mathbb{E}[\left\|g_{k}^t-\nabla\mathcal{L}_k(\omega_{k}^t)\right\|_2^2]\leqslant\sigma^2,\forall k\in\{1,2,\ldots,m\}.
    \label{eq:bounded variance}
\end{equation}
\textbf{Assumption 3.} (Bounded Gradient). The stochastic gradient of the global prototype is bounded by $G$:
\begin{equation}
\scalebox{0.9}{$
        \mathbb{E}[\|\nabla\mathcal{L}_{server}\|_2]\leqslant G
    \label{eq:bounded eulidean norm}
$}
\end{equation}
We denote $e \in \{0, 1, 2, \ldots, E\}$ as the local iteration, and t as the global round. Here, $tE$ indicates the time step before orthogonality regularization at $t$ round, and $tE + 0$ represents the interval between global prototype training and the first iteration of the $(t+1)$ round. Under above assumptions, we present the theoretical results at the non-convex condition. Since all client-side loss function share the same property, we omit the subscript $k$ and denote the loss function as $\mathcal{L}$. The detailed proof process is in the Appendix.\\
\textbf{Lemma 1.} Based on Assumption 1 and 2, after clients training one round, the local client model loss satisfies \cite{tan2022fedproto}:
\begin{equation}
    \begin{aligned}
        \mathbb{E}\left[\mathcal{L}^{(t+1)E}\right] \leqslant &\mathcal{L}^{tE+0} - \left(\eta - \frac{L_1 \eta^2}{2}\right) \sum_{e=0}^{E-1} \|\mathcal{L}^{tE+e}\|_2^2 + \\
        &\frac{L_1 E \eta^2}{2} \sigma^2
    \end{aligned}
\label{eq:lemma 4.1}
\end{equation}
\textbf{Lemma 2.} Based on Assumption 3, we further analyze how the introduction of global prototype orthogonality regularization influences the client-side loss:
\begin{equation}
\mathbb{E}\left[\mathcal{L}^{(t+1)E+0}\right]\leqslant\mathbb{E}\left[\mathcal{L}^{(t+1)E}\right]+\lambda_c \eta G.
    \label{eq:lemma 4.2}
\end{equation}
\textbf{Theorem 1.} Leveraging Lemma 1 and Lemma 2, we establish our main theoretical convergence results of the loss reduction over multiple steps in local training:
\begin{equation}
    \begin{aligned}
        \mathbb{E}\left[\mathcal{L}^{(t+1)E+0}\right]\leqslant &\mathcal{L}^{tE+0}-\left(\eta-\frac{L_1\eta^2}2\right)\sum_{e=0}^{E-1}\|\mathcal{L}^{tE+e}\|_2^2 +\\
        &\frac{\eta^2 L_1E\sigma^2}2 +\lambda_c \eta G.
    \end{aligned}
    \label{eq:theroem 1}
\end{equation}
\textbf{Theorem 2.} We extend the analysis to multiple rounds, providing a convergence rate for FedORGP in a non-convex setting. For an arbitrary client and any $\epsilon> 0$, the following inequality holds:
\begin{equation}
    \begin{aligned}
        \frac1T\sum_{t=0}^{T-1}\sum_{e=0}^{E-1}\mathbb{E}\left[\|\mathcal{L}^{tE+e}\|_2^2\right]\leqslant &\frac{2\left(\mathcal{L}^{t=0}-\mathcal{L}^*\right)}{T\eta\left(2-L_1\eta\right)}+\frac{L_1E\eta\sigma^2}{2-L_1\eta}+\\
        &\frac{2\lambda_c G}{2-L_1\eta}\leqslant \epsilon\\
    \end{aligned}
    \label{eq:theroem 2}
\end{equation}
when $\eta<\frac{2(\epsilon-\lambda_c G)}{L_1\left(\epsilon+E\sigma^2\right)}$ and $\lambda_c < \frac{\epsilon}{G}$.
 It shows that the model can converge to a solution with a rate that is proportional to $O(1/T)$, where $T$ is the number of communication rounds.

%% file: sec/Experiments.tex
\section{Experiments}
\label{sec:experiments}

\begin{table*}[t]
\caption{The test accuracy (\%) on four datasets in the pathological and practical settings using the HMG$_8$ model group.}
\resizebox{\textwidth}{!}{%
\begin{tabular}{|l|cccc|cccc|}
\hline
Settings  & \multicolumn{4}{c|}{Pathological Setting}                                                                                                                                & \multicolumn{4}{c|}{Practical Setting}                                                                                                                                   \\ \hline
Datasets  & \multicolumn{1}{c|}{Cifar-10}               & \multicolumn{1}{c|}{Cifar-100}              & \multicolumn{1}{c|}{Flowers102}            & \multicolumn{1}{l|}{TinyImagenet} & \multicolumn{1}{c|}{Cifar-10}               & \multicolumn{1}{c|}{Cifar-100}              & \multicolumn{1}{c|}{Flowers102}            & \multicolumn{1}{l|}{TinyImagenet} \\ \hline
FML       & \multicolumn{1}{c|}{82.56 ± 0.14}          & \multicolumn{1}{c|}{50.27 ± 0.09}          & \multicolumn{1}{c|}{52.17 ± 0.98}          & 33.37 ± 0.34                      & \multicolumn{1}{c|}{91.85 ± 0.24}          & \multicolumn{1}{c|}{46.45 ± 0.17}          & \multicolumn{1}{c|}{48.51 ± 0.57}          & 37.67 ± 0.19                      \\ \hline
LG-FedAvg & \multicolumn{1}{c|}{83.63 ± 0.09}          & \multicolumn{1}{c|}{55.51 ± 0.10}          & \multicolumn{1}{c|}{58.90 ± 0.22}          & 34.99 ± 0.29                      & \multicolumn{1}{c|}{92.34 ± 0.06}          & \multicolumn{1}{c|}{49.82 ± 0.16}          & \multicolumn{1}{c|}{53.65 ± 0.46}          & 38.35 ± 0.27                      \\ \hline
FedGen    & \multicolumn{1}{c|}{83.63 ± 0.24}          & \multicolumn{1}{c|}{55.37 ± 0.35}          & \multicolumn{1}{c|}{59.39 ± 0.19}          & 35.18 ± 0.12                      & \multicolumn{1}{c|}{92.43 ± 0.11}          & \multicolumn{1}{c|}{49.90 ± 0.21}          & \multicolumn{1}{c|}{54.59 ± 0.34}          & 38.44 ± 0.14                      \\ \hline
FedProto  & \multicolumn{1}{c|}{80.63 ± 0.05}          & \multicolumn{1}{c|}{51.89 ± 0.31}          & \multicolumn{1}{c|}{54.48 ± 0.20}          & 33.78 ± 0.20                      & \multicolumn{1}{c|}{79.32 ± 0.22}          & \multicolumn{1}{c|}{42.77 ± 0.13}          & \multicolumn{1}{c|}{26.66 ± 0.57}          & 24.57 ± 0.08                      \\ \hline
FedKD     & \multicolumn{1}{c|}{83.41 ± 0.66}          & \multicolumn{1}{c|}{53.01 ± 1.48}          & \multicolumn{1}{c|}{52.61 ± 0.74}          & 34.70 ± 0.91                      & \multicolumn{1}{c|}{91.85 ± 0.31}          & \multicolumn{1}{c|}{48.95 ± 1.24}          & \multicolumn{1}{c|}{51.44 ± 0.75}          & 39.13 ± 0.36                      \\ \hline
FedGH     & \multicolumn{1}{c|}{83.49 ± 0.29}          & \multicolumn{1}{c|}{55.25 ± 0.12}          & \multicolumn{1}{c|}{60.15 ± 1.08}          & 35.04 ± 0.16                      & \multicolumn{1}{c|}{92.66 ± 0.10}          & \multicolumn{1}{c|}{49.52 ± 0.05}          & \multicolumn{1}{c|}{54.24 ± 0.50}          & 38.67 ± 0.11                      \\ \hline
FedTGP    & \multicolumn{1}{c|}{84.75 ± 0.06}          & \multicolumn{1}{c|}{53.12 ± 0.24}          & \multicolumn{1}{c|}{58.60 ± 0.14}          & 32.41 ± 0.09                      & \multicolumn{1}{c|}{92.26 ± 0.68}          & \multicolumn{1}{c|}{49.09 ± 0.16}          & \multicolumn{1}{c|}{56.62 ± 0.40}          & 37.14 ± 0.16                      \\ \hline
FedORGP   & \multicolumn{1}{c|}{\textbf{85.71 ± 0.10}} & \multicolumn{1}{c|}{\textbf{61.12 ± 0.15}} & \multicolumn{1}{c|}{\textbf{64.15 ± 1.66}} & \textbf{37.04 ± 0.56}             & \multicolumn{1}{c|}{\textbf{94.01 ± 0.03}} & \multicolumn{1}{c|}{\textbf{55.33 ± 0.23}} & \multicolumn{1}{c|}{\textbf{60.34 ± 0.48}} & \textbf{40.15 ± 0.40}             \\ \hline
\end{tabular}
}
\label{tab:test all comparison_results} 
\end{table*}

\subsection{Setup}
\textbf{Datasets.} To evaluate our model, we conduct experiments on four image classification datasets: CIFAR-10, CIFAR-100 \cite{Cifar}, Flowers102 \cite{flowers102} and TinyImagenet \cite{tinyimagenet}.\\
\textbf{Baselines.} To evaluate our proposed FedORGP, we compare it with seven popular methods that are applicable in HtFL, including LG-FedAvg \cite{lgfedavg}, FML \cite{fml}, FedGen \cite{FedGen}, FedKD \cite{fedkd}, FedProto \cite{tan2022fedproto}, FedGH \cite{yi2023fedgh} and FedTGP \cite{fedtgp}.\\
\textbf{Model heterogeneity.} To comprehensively evaluate the robustness and adaptability of our algorithm across different model architectures, we test our approach on four heterogeneous model groups (HMG): HMG$_3$ (4-layer CNN \cite{fedavg}, GoogleNet \cite{googlenet} and MobileNet\_v2 \cite{mobilenet}), HMG$_5$ (ResNet18/34/50/101/152 \cite{resnet}), HMG$_8$ (Combines HMG$_3$ and HMG$_5$), and HMG$_{10}$ (Extends HMG$_8$ with DenseNet121 \cite{desnet} and EfficientNet-B0 \cite{tan2019efficientnet}) \cite{fedtgp}. These HMGs span lightweight and complex convolutional networks, enabling evaluation of the proposed method across a broad range of model complexities.\\
\textbf{Statistical heterogeneity.} We conduct extensive experiments with two widely used statistically heterogeneous settings, the pathological setting and the practical setting \cite{fedavg, FedGen, FedDF}. For the pathological setting, we assign a fixed number of classes to the client. For the practical setting, we leverage Dirichlet distribution to simulate more realistic class imbalance across clients. The hyperparameter $\alpha$ controls the strength of heterogeneity. Notably, a smaller $\alpha$ implies a higher non-IID data distribution among clients.\\
\textbf{Training configuration.} Our experiments are conducted on an x86\_64 architecture with Ubuntu as the operating system. The platform is equipped with 8 NVIDIA V100 GPUs, each with 32 GB of memory, and CUDA version 12.2. Unless explicitly specified, we use the following settings. The federated learning setup includes a default configuration of 20 clients and the client participation ratio $\rho$ = 1. Both the client and server employ the SGD optimizer with a learning rate of 0.01, and both undergo training for a single epoch. The batch sizes for client $B$ and server $B_p$ are both set to 32. For model heterogeneity, we use the HMG$_8$ by default. For statistical heterogeneity testing, under the pathological setting, we distribute unbalanced data of 2/10/10/20 classes to each client from a total of 10/100/102/200 classes on Cifar-10/Cifar-100/Flowers102/TinyImagenet datasets. In the practical setting, a Dirichlet distribution with $\alpha$ = 0.05 is used to simulate real life scenario. Each client's private data is divided into a training set (75\%) and a test set (25\%). Each algorithm is trained three times, with each training run consisting of 100 epochs. For the evaluation, we count the best accuracy achieved in each training run, and the final result is calculated as the mean and variance of the accuracy of the three runs.\\
\textbf{Evaluation Metrics.} We measure the average test accuracy (\%) across all client models. We further evaluate the robustness of our algorithm under varying client participation rates, degrees of non-IID data, and feature dimension $K$.

\subsection{Performance Comparison}
The test accuracy of all methods across four datasets is presented in Tab. \ref{tab:test all comparison_results}. FedORGP consistently outperforms all baselines on these datasets, achieving up to a 8\% gain over FedTGP in pathological settings and 6.24\% in practical settings on Cifar-100. This improvement is due to the fact that FedORGP can separate features in terms of angles, which can be well integrated with the CE loss, thereby improving classification results more effectively than methods which only increase the Euclidean distance to perform prototype separation in high-dimensional space. In addition, FedORGP is also more efficient. While FedTGP requires 100 epochs of training on the server, FedORGP requires only 1 epoch of training with orthogonal regularisation.

\begin{table}[ht]
\caption{Test on Cifar-100 (10 classes per client) with three different HMGs to test robustness to heterogeneous models.}
\label{tab:test HMG} 
\resizebox{\linewidth}{!}{%
\begin{tabular}{|l|ccc|}
\hline
          & \multicolumn{3}{c|}{Heterogenous Model Groups}                                                                  \\ \hline
Settings  & \multicolumn{1}{c|}{HMG$_3$}               & \multicolumn{1}{c|}{HMG$_5$}               & HMG$_{10}$            \\ \hline
FML       & \multicolumn{1}{c|}{58.72 ± 0.15}          & \multicolumn{1}{c|}{42.84 ± 0.41}          & 45.50 ± 0.50          \\ \hline
LG-FedAvg & \multicolumn{1}{c|}{59.62 ± 0.24}          & \multicolumn{1}{c|}{51.46 ± 0.21}          & 50.19 ± 0.19          \\ \hline
FedGen    & \multicolumn{1}{c|}{59.82 ± 0.38}          & \multicolumn{1}{c|}{50.97 ± 0.26}          & 50.07 ± 0.13          \\ \hline
FedProto  & \multicolumn{1}{c|}{58.83 ± 0.52}          & \multicolumn{1}{c|}{49.08 ± 0.00}          & 48.04 ± 0.02          \\ \hline
FedKD     & \multicolumn{1}{c|}{\textbf{61.05 ± 0.04}} & \multicolumn{1}{c|}{46.24 ± 2.55}          & 47.57 ± 2.00          \\ \hline
FedGH     & \multicolumn{1}{c|}{59.21 ± 0.15}          & \multicolumn{1}{c|}{50.76 ± 0.30}          & 50.41 ± 0.19          \\ \hline
FedTGP    & \multicolumn{1}{c|}{60.41 ± 0.56}          & \multicolumn{1}{c|}{48.24 ± 0.47}          & 48.11 ± 0.07          \\ \hline
FedORGP   & \multicolumn{1}{c|}{60.64 ± 0.50}          & \multicolumn{1}{c|}{\textbf{61.58 ± 0.42}} & \textbf{58.54 ± 0.65} \\ \hline
\end{tabular}
}
\end{table}

\subsection{Impact of Model Heterogeneity}
To examine the impact of model heterogeneity in HtFL, we assess the performance of FedORGP on three additional HMG settings. We show results in Tab. \ref{tab:test HMG}, our findings indicate that all baselines perform well under the HMG$_3$ condition, as the reduced model heterogeneity significantly lessened its adverse impact on performance, while all methods perform worse with larger model heterogeneity. As the degree of model heterogeneity rises, the advantages of FedORGP becomes more pronounced, up to 10.12\% at HMG$_5$ and 8.13\% at HMG$_{10}$. In addition, FedORGP also has the smallest performance reduction when model heterogeneity increases, which shows that FedORGP is robust to model heterogeneity. 

\subsection{Robustness to Participation Rate}
To compare FedORGP against baselines under varying total client numbers $M$ and client participation rates $\rho$, we design three distinct settings. The results, shown in Tab. \ref{tab:pariticipation ratio}, ensure a consistent number of participating clients per round. FedORGP aims to promote directional independence, which is seamlessly integrated with CE loss. This separation minimizes inter-class overlap, thereby enabling the model to achieve robust generalization across clients with varying participation rates.

\begin{table}[ht]
\caption{Test on Cifar-100 (10 classes per client) under three participation rate settings with HMG$_8$. “-” means the baseline can't converge.}
\label{tab:pariticipation ratio} 
\resizebox{\linewidth}{!}{%
\begin{tabular}{|l|ccc|}
\hline
          & \multicolumn{3}{c|}{Different $M$ Clients and Join Ratio}                                                       \\ \hline
Settings  & \multicolumn{1}{c|}{$M$=40, $\rho$=50\%}   & \multicolumn{1}{c|}{$M$=80, $\rho$=25\%}   & $M$=100, $\rho$=20\%  \\ \hline
FML       & \multicolumn{1}{c|}{33.96 ± 0.25}          & \multicolumn{1}{c|}{19.72 ± 0.11}          & 30.65 ± 0.25          \\ \hline
LG-FedAvg & \multicolumn{1}{c|}{47.05 ± 0.09}          & \multicolumn{1}{c|}{39.59 ± 0.23}          & 40.98 ± 0.51          \\ \hline
FedGen    & \multicolumn{1}{c|}{47.05 ± 0.40}          & \multicolumn{1}{c|}{39.76 ± 0.35}          & 41.10 ± 0.53          \\ \hline
FedProto  & \multicolumn{1}{c|}{38.30 ± 0.53}          & \multicolumn{1}{c|}{13.49 ± 0.84}          & 16.16 ± 0.55          \\ \hline
FedKD     & \multicolumn{1}{c|}{38.12 ± 3.36}          & \multicolumn{1}{c|}{21.76 ± 1.60}          & 34.59 ± 2.73          \\ \hline
FedGH     & \multicolumn{1}{c|}{46.82 ± 0.24}          & \multicolumn{1}{c|}{40.08 ± 0.35}          & -                     \\ \hline
FedTGP    & \multicolumn{1}{c|}{44.87 ± 0.58}          & \multicolumn{1}{c|}{37.51 ± 0.36}          & 36.75 ± 0.42          \\ \hline
FedORGP   & \multicolumn{1}{c|}{\textbf{54.88 ± 0.45}} & \multicolumn{1}{c|}{\textbf{44.86 ± 0.11}} & \textbf{43.22 ± 0.45} \\ \hline
\end{tabular}
}
\end{table}



\begin{table}[ht]
\caption{Test on two data distributions under various degrees of non-IID on the Cifar-100 dataset with HMG$_8$.}
\label{tab:noniid} 
\resizebox{\linewidth}{!}{%
\begin{tabular}{|l|cc|cc|}
\hline
Settings  & \multicolumn{2}{c|}{Pathological Setting}                          & \multicolumn{2}{c|}{Practical Setting}                             \\ \hline
          & \multicolumn{1}{c|}{5 classes/client}      & 20 classes/client     & \multicolumn{1}{c|}{$\alpha$=0.1}          & $\alpha$=0.01         \\ \hline
FML       & \multicolumn{1}{c|}{66.40 ± 0.33}          & 33.13 ± 0.24          & \multicolumn{1}{c|}{37.30 ± 0.09}          & 61.17 ± 0.24          \\ \hline
LG-FedAvg & \multicolumn{1}{c|}{71.84 ± 0.17}          & 37.22 ± 0.36          & \multicolumn{1}{c|}{39.97 ± 0.11}          & 66.06 ± 0.14          \\ \hline
FedGen    & \multicolumn{1}{c|}{71.68 ± 0.09}          & 37.47 ± 0.09          & \multicolumn{1}{c|}{40.33 ± 0.37}          & 66.06 ± 0.08          \\ \hline
FedProto  & \multicolumn{1}{c|}{70.27 ± 0.07}          & 35.15 ± 0.18          & \multicolumn{1}{c|}{34.30 ± 0.16}          & 56.97 ± 0.36          \\ \hline
FedKD     & \multicolumn{1}{c|}{69.75 ± 1.35}          & 34.24 ± 0.79          & \multicolumn{1}{c|}{39.81 ± 1.12}          & 63.86 ± 1.46          \\ \hline
FedGH     & \multicolumn{1}{c|}{71.42 ± 0.29}          & 36.99 ± 0.23          & \multicolumn{1}{c|}{39.90 ± 0.21}          & 65.90 ± 0.27          \\ \hline
FedTGP    & \multicolumn{1}{c|}{69.86 ± 0.33}          & 35.98 ± 0.30          & \multicolumn{1}{c|}{38.60 ± 0.10}          & 67.12 ± 0.11          \\ \hline
FedORGP   & \multicolumn{1}{c|}{\textbf{75.95 ± 0.24}} & \textbf{45.45 ± 0.34} & \multicolumn{1}{c|}{\textbf{44.70 ± 0.29}} & \textbf{70.32 ± 0.36} \\ \hline
\end{tabular}
}
\end{table}

\subsection{Robustness to Statistical Heterogeneity}
To evaluate the robustness of all baseline methods under different non-IID data distributions, we conducted experiments on two dataset configurations. As demonstrated in Tab. \ref{tab:noniid}, FedORGP ensures optimal inter-class separation by leveraging orthogonality regularization, reducing embeddings overlap even in highly skewed data distributions. This demonstrates its effectiveness in promoting generalization in statistical heterogeneity environments.

\subsection{Impact of Feature Dimension}
\begin{figure}[ht]
    \centering
    \begin{subfigure}[b]{0.49\linewidth}
        \centering
        \includegraphics[width=\linewidth]{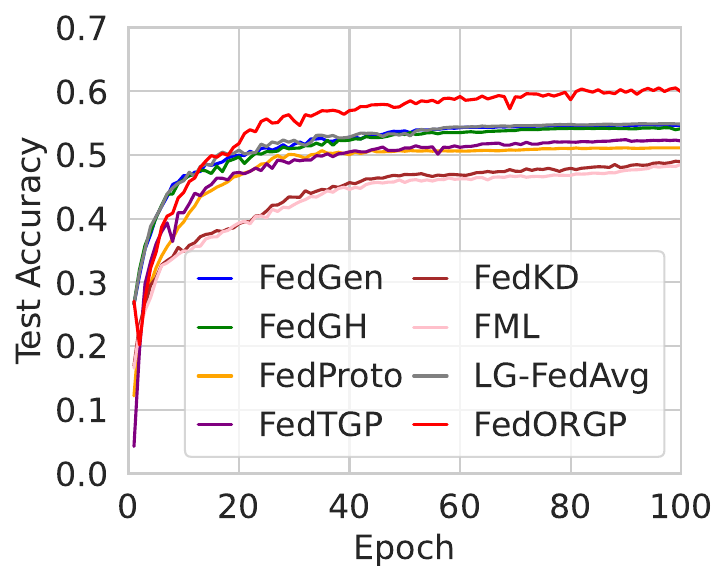}
        \caption{K=128}
        \label{fig:K=128}
    \end{subfigure}
    \hfill
    \begin{subfigure}[b]{0.49\linewidth}
        \centering
        \includegraphics[width=\linewidth]{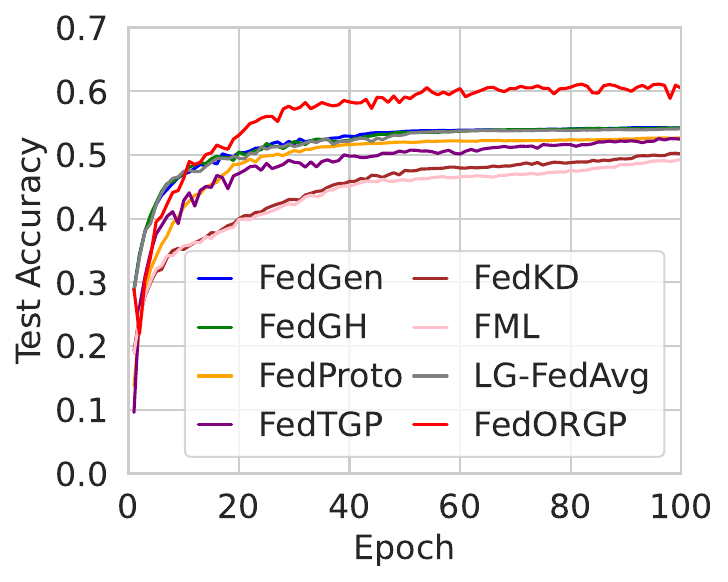}
        \caption{K=256}
        \label{fig:K=256}
    \end{subfigure}
    \hfill
    \begin{subfigure}[b]{0.49\linewidth}
        \centering
        \includegraphics[width=\linewidth]{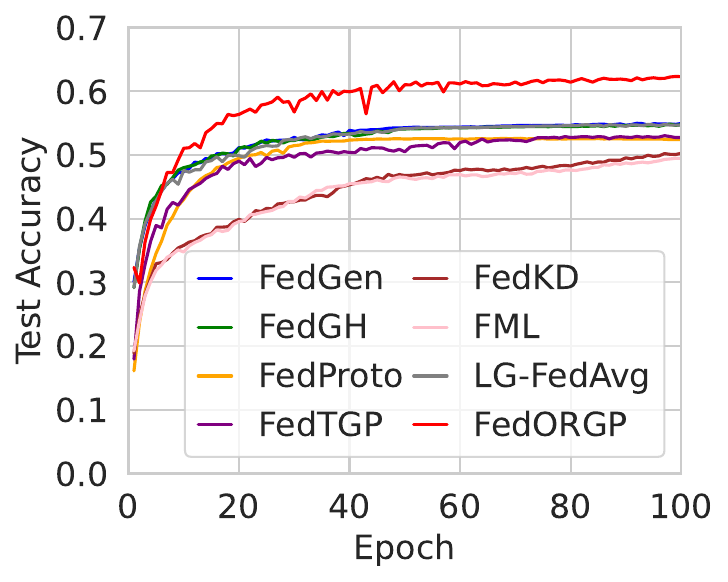}
        \caption{K=512}
        \label{fig:K=512}
    \end{subfigure}
    \hfill
    \begin{subfigure}[b]{0.49\linewidth}
        \centering
        \includegraphics[width=\linewidth]{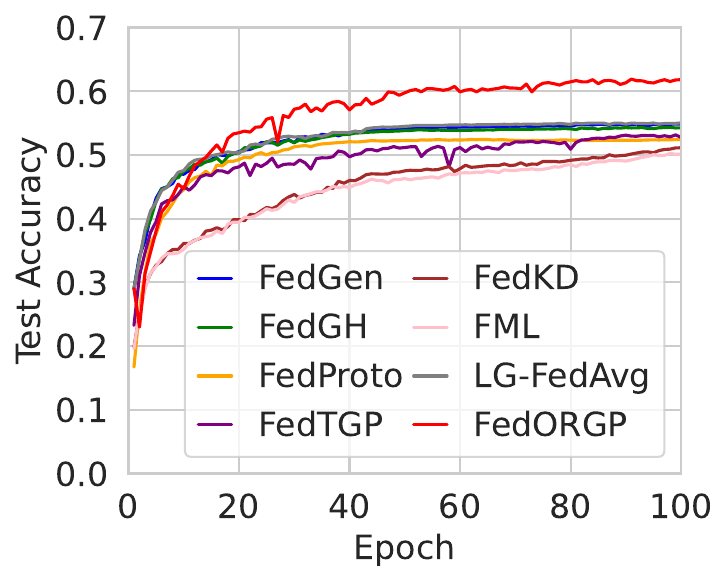}
        \caption{K=1024}
        \label{fig:K=1024}
    \end{subfigure}
    \caption{Test on Cifar-100 (10 classes per client) under different feature dimensions with HMG$_8$.}
    \label{fig:k-dimension}
\end{figure}

To assess the robustness of all baselines to the $K$, we performe experiments with three feature dimensions: 128, 256, and 1024, as indicated in Fig. \ref{fig:k-dimension}. Results indicate that all baselines maintain a high degree of robustness, showing minimal variation in performance.

\subsection{Ablation Study}
This section analyzes the ablation study in Tab. \ref{tab:ablation}. FedORGP without orthogonality regularization (FedORGP w/o OC) aligns the embedding layer with the global prototype derived from weighted averaging. While the performance on CIFAR-10 is passable, FedORGP w/o OC exhibits a sharp performance decline on CIFAR-100 and Flowers102 datasets, likely due to substantial global prototype overlap. Introducing the orthogonality regularization in FedORGP markedly enhances prototype separation, improving performance by explicitly increasing angular separation across classes. FedORGP outperforms FedTGP, as reducing prototype similarity integrates more effectively with cross-entropy (CE) loss than merely increasing prototype distances.

\begin{table}[ht]
\caption{Test on three datasets under the practical setting using the HMG$_8$ for ablation study.}
\label{tab:ablation} 
\resizebox{\linewidth}{!}{%
\begin{tabular}{|l|l|l|l|}
\hline
           & \multicolumn{1}{c|}{FedTGP}     & \multicolumn{1}{c|}{FedORGP w/o OR} & \multicolumn{1}{c|}{FedORGP}               \\ \hline
Cifar-10    & 84.75 ± 0.06 & 78.05 ± 0.52                          & \textbf{85.71 ± 0.10} \\ \hline
Cifar-100   & 53.12 ± 0.24 & 35.11 ± 0.63                          & \textbf{61.12 ± 0.15} \\ \hline
Flowers102 & 58.60 ± 0.14 & 42.43 ± 0.57                          & \textbf{64.15 ± 1.66} \\ \hline
\end{tabular}
 }
\end{table}

\subsection{Hyperparameter Combination}
We conducted a grid search across $(\lambda_c, \lambda_s, \gamma)$ in Tab. \ref{tab:hyperpara} to determine the optimal combination of hyperparameters. The selected values for $\lambda_s$ and $\gamma$, specifically (1, 10), enable the orthogonality constraint to effectively enhance inter-class separation without sacrificing the integrity of class representations. This configuration fosters a more robust global prototype compared to the (10, 100) pairing at the same ratio. Additionally, a higher $\lambda_c$ imposes a stricter alignment constraint, thereby improving feature representation consistency across clients, which is crucial for maintaining a unified representation space against statistical heterogeneity. The highest accuracy 61.15\% is achieved with the (100, 1, 10) setting.

\begin{table}[ht]
\caption{Hyperparameter search on Cifar-100 with HMG$_8$.}
\label{tab:hyperpara} 
\resizebox{\linewidth}{!}{%
\begin{tabular}{|cccc|cccc|cccc|}
\toprule
\huge $\lambda_c$ & \huge $\lambda_s$ &\huge  $\gamma$ &\huge  Acc &\huge  $\lambda_c$ & \huge $\lambda_s$ & \huge $\gamma$ & \huge Acc & \huge $\lambda_c$ & \huge $\lambda_s$ & \huge $\gamma$ & \huge Acc \\
\midrule
\huge 100.0 & \huge  100.0 & \huge  100.0 & \huge  40.34 & \huge  10.0 & \huge  100.0 & \huge  100.0 & \huge  50.67 & \huge  1.0 & \huge  100.0 & \huge  100.0 & \huge  53.94 \\
\huge 100.0 & \huge  100.0 & \huge  10.0 & \huge  40.32 & \huge  10.0 & \huge  100.0 & \huge  10.0 & \huge  48.12 & \huge  1.0 & \huge  100.0 & \huge  10.0 & \huge  54.43 \\
\huge 100.0 & \huge  100.0 & \huge  1.0 & \huge  48.32 & \huge  10.0 & \huge  100.0 & \huge  1.0 & \huge  51.52 & \huge  1.0 & \huge  100.0 & \huge  1.0 & \huge  54.82 \\
\huge 100.0 & \huge  10.0 & \huge  100.0 & \huge  59.17 & \huge  10.0 & \huge  10.0 & \huge  100.0 & \huge  56.60 & \huge  1.0 & \huge  10.0 & \huge  100.0 & \huge  53.94 \\
\huge 100.0 & \huge  10.0 & \huge  10.0 & \huge  55.37 & \huge  10.0 & \huge  10.0 & \huge  10.0 & \huge  54.76 & \huge  1.0 & \huge  10.0 & \huge  10.0 & \huge  54.34 \\
\huge 100.0 & \huge  10.0 & \huge  1.0 & \huge  51.96 & \huge  10.0 & \huge  10.0 & \huge  1.0 & \huge  53.47 & \huge  1.0 & \huge  10.0 & \huge  1.0 & \huge  54.22 \\
\huge 100.0 & \huge  1.0 & \huge  100.0 & \huge  58.97 & \huge  10.0 & \huge  1.0 & \huge  100.0 & \huge  55.14 & \huge  1.0 & \huge  1.0 & \huge  100.0 & \huge  50.64 \\
\huge \textbf{100.0} & \huge  \textbf{1.0} & \huge  \textbf{10.0} & \huge  \textbf{61.15} & \huge  10.0 & \huge  1.0 & \huge  10.0 & \huge  57.35 & \huge  1.0 & \huge  1.0 & \huge  10.0 & \huge  54.94 \\
\huge 100.0 & \huge  1.0 & \huge  1.0 & \huge  59.51 & \huge  10.0 & \huge  1.0 & \huge  1.0 & \huge  57.11 & \huge  1.0 & \huge  1.0 & \huge  1.0 & \huge  54.20 \\
\bottomrule
\end{tabular}
}
\end{table}

%% file: sec/Conclusion.tex
\section{Conclusion}
\label{sec:conclusion}

In this work, we propose a novel HtFL method, FedORGP, which employs orthogonality regularization on global prototypes at the server side to improve inter-prototype separation and maximize directional independence while preserving semantic integrity. On the client side, FedORGP aligns the embeddings with prototypes, enhancing its compatibility with CE loss. This approach overcomes the limitations of FedTGP by enhancing prototype separation through angular separation rather than merely increasing Euclidean distances between prototypes. Extensive experiments demonstrate that FedORGP consistently outperforms all baseline methods, and exhibits superior performance in scenarios where model and statistical heterogeneity coexist.

%% file: main.bbl
\begin{thebibliography}{45}
\providecommand{\natexlab}[1]{#1}
\providecommand{\url}[1]{\texttt{#1}}
\expandafter\ifx\csname urlstyle\endcsname\relax
  \providecommand{\doi}[1]{doi: #1}\else
  \providecommand{\doi}{doi: \begingroup \urlstyle{rm}\Url}\fi

\bibitem[Aouedi et~al.(2022)Aouedi, Sacco, Piamrat, and Marchetto]{uncentralized}
Ons Aouedi, Alessio Sacco, Kandaraj Piamrat, and Guido Marchetto.
\newblock Handling privacy-sensitive medical data with federated learning: challenges and future directions.
\newblock \emph{IEEE journal of biomedical and health informatics}, 27\penalty0 (2):\penalty0 790--803, 2022.

\bibitem[Bakman et~al.(2023)Bakman, Yaldiz, Ezzeldin, and Avestimehr]{FOT}
Yavuz~Faruk Bakman, Duygu~Nur Yaldiz, Yahya~H Ezzeldin, and Salman Avestimehr.
\newblock Federated orthogonal training: Mitigating global catastrophic forgetting in continual federated learning.
\newblock \emph{arXiv preprint arXiv:2309.01289}, 2023.

\bibitem[Chrabaszcz et~al.(2017)Chrabaszcz, Loshchilov, and Hutter]{tinyimagenet}
Patryk Chrabaszcz, Ilya Loshchilov, and Frank Hutter.
\newblock A downsampled variant of imagenet as an alternative to the cifar datasets.
\newblock \emph{arXiv preprint arXiv:1707.08819}, 2017.

\bibitem[Dai et~al.(2023)Dai, Chen, Li, Heinecke, Sun, and Xu]{FedNH}
Yutong Dai, Zeyuan Chen, Junnan Li, Shelby Heinecke, Lichao Sun, and Ran Xu.
\newblock Tackling data heterogeneity in federated learning with class prototypes.
\newblock In \emph{Proceedings of the AAAI Conference on Artificial Intelligence}, pages 7314--7322, 2023.

\bibitem[Diao et~al.(2020)Diao, Ding, and Tarokh]{diao2020heterofl}
Enmao Diao, Jie Ding, and Vahid Tarokh.
\newblock Heterofl: Computation and communication efficient federated learning for heterogeneous clients.
\newblock \emph{arXiv preprint arXiv:2010.01264}, 2020.

\bibitem[Gao et~al.(2023)Gao, Li, Lu, and Wu]{Fedios}
Lingzhi Gao, Zexi Li, Yang Lu, and Chao Wu.
\newblock Fedios: Decoupling orthogonal subspaces for personalization in feature-skew federated learning.
\newblock \emph{arXiv preprint arXiv:2311.18559}, 2023.

\bibitem[Gupta and Alam(2022)]{promising}
Ruchi Gupta and Tanweer Alam.
\newblock Survey on federated-learning approaches in distributed environment.
\newblock \emph{Wireless personal communications}, 125\penalty0 (2):\penalty0 1631--1652, 2022.

\bibitem[He et~al.(2016)He, Zhang, Ren, and Sun]{resnet}
Kaiming He, Xiangyu Zhang, Shaoqing Ren, and Jian Sun.
\newblock Deep residual learning for image recognition.
\newblock In \emph{Proceedings of the IEEE conference on computer vision and pattern recognition}, pages 770--778, 2016.

\bibitem[Hersche et~al.(2022)Hersche, Karunaratne, Cherubini, Benini, Sebastian, and Rahimi]{C-FSCIL}
Michael Hersche, Geethan Karunaratne, Giovanni Cherubini, Luca Benini, Abu Sebastian, and Abbas Rahimi.
\newblock Constrained few-shot class-incremental learning.
\newblock In \emph{Proceedings of the IEEE/CVF conference on computer vision and pattern recognition}, pages 9057--9067, 2022.

\bibitem[Horvath et~al.(2021)Horvath, Laskaridis, Almeida, Leontiadis, Venieris, and Lane]{horvath2021fjord}
Samuel Horvath, Stefanos Laskaridis, Mario Almeida, Ilias Leontiadis, Stylianos Venieris, and Nicholas Lane.
\newblock Fjord: Fair and accurate federated learning under heterogeneous targets with ordered dropout.
\newblock \emph{Advances in Neural Information Processing Systems}, 34:\penalty0 12876--12889, 2021.

\bibitem[Huang et~al.(2017)Huang, Liu, Van Der~Maaten, and Weinberger]{desnet}
Gao Huang, Zhuang Liu, Laurens Van Der~Maaten, and Kilian~Q Weinberger.
\newblock Densely connected convolutional networks.
\newblock In \emph{Proceedings of the IEEE conference on computer vision and pattern recognition}, pages 4700--4708, 2017.

\bibitem[Huang et~al.(2023)Huang, Ye, Shi, Li, and Du]{FPL}
Wenke Huang, Mang Ye, Zekun Shi, He Li, and Bo Du.
\newblock Rethinking federated learning with domain shift: A prototype view.
\newblock In \emph{2023 IEEE/CVF Conference on Computer Vision and Pattern Recognition (CVPR)}, pages 16312--16322. IEEE, 2023.

\bibitem[Kairouz et~al.(2021)Kairouz, McMahan, Avent, Bellet, Bennis, Bhagoji, Bonawitz, Charles, Cormode, Cummings, et~al.]{FLadvance}
Peter Kairouz, H~Brendan McMahan, Brendan Avent, Aur{\'e}lien Bellet, Mehdi Bennis, Arjun~Nitin Bhagoji, Kallista Bonawitz, Zachary Charles, Graham Cormode, Rachel Cummings, et~al.
\newblock Advances and open problems in federated learning.
\newblock \emph{Foundations and trends{\textregistered} in machine learning}, 14\penalty0 (1--2):\penalty0 1--210, 2021.

\bibitem[Krizhevsky et~al.(2009)Krizhevsky, Hinton, et~al.]{Cifar}
Alex Krizhevsky, Geoffrey Hinton, et~al.
\newblock Learning multiple layers of features from tiny images.
\newblock \emph{Technical Report}, 2009.

\bibitem[Lee et~al.(2024)Lee, Jeong, Kim, Oh, and Yun]{FedSOL}
Gihun Lee, Minchan Jeong, Sangmook Kim, Jaehoon Oh, and Se-Young Yun.
\newblock Fedsol: Stabilized orthogonal learning with proximal restrictions in federated learning.
\newblock In \emph{Proceedings of the IEEE/CVF Conference on Computer Vision and Pattern Recognition}, pages 12512--12522, 2024.

\bibitem[Li and Wang(2019)]{li2019fedmd}
Daliang Li and Junpu Wang.
\newblock Fedmd: Heterogenous federated learning via model distillation.
\newblock \emph{arXiv preprint arXiv:1910.03581}, 2019.

\bibitem[Li et~al.(2020{\natexlab{a}})Li, Sahu, Talwalkar, and Smith]{fl_challenges_methods_directions}
Tian Li, Anit~Kumar Sahu, Ameet Talwalkar, and Virginia Smith.
\newblock Federated learning: Challenges, methods, and future directions.
\newblock \emph{IEEE signal processing magazine}, 37\penalty0 (3):\penalty0 50--60, 2020{\natexlab{a}}.

\bibitem[Li et~al.(2020{\natexlab{b}})Li, Sahu, Zaheer, Sanjabi, Talwalkar, and Smith]{FedProx}
Tian Li, Anit~Kumar Sahu, Manzil Zaheer, Maziar Sanjabi, Ameet Talwalkar, and Virginia Smith.
\newblock Federated optimization in heterogeneous networks.
\newblock In \emph{Proceedings of Machine Learning and Systems}, pages 429--450, 2020{\natexlab{b}}.

\bibitem[Li et~al.(2023)Li, Shang, He, Lin, and Wu]{nofearofclassifierbiases}
Zexi Li, Xinyi Shang, Rui He, Tao Lin, and Chao Wu.
\newblock No fear of classifier biases: Neural collapse inspired federated learning with synthetic and fixed classifier.
\newblock In \emph{Proceedings of the IEEE/CVF International Conference on Computer Vision}, pages 5319--5329, 2023.

\bibitem[Liang et~al.(2020)Liang, Liu, Ziyin, Allen, Auerbach, Brent, Salakhutdinov, and Morency]{lgfedavg}
Paul~Pu Liang, Terrance Liu, Liu Ziyin, Nicholas~B Allen, Randy~P Auerbach, David Brent, Ruslan Salakhutdinov, and Louis-Philippe Morency.
\newblock Think locally, act globally: Federated learning with local and global representations.
\newblock \emph{arXiv preprint arXiv:2001.01523}, 2020.

\bibitem[Lin et~al.(2020)Lin, Kong, Stich, and Jaggi]{FedDF}
Tao Lin, Lingjing Kong, Sebastian~U Stich, and Martin Jaggi.
\newblock Ensemble distillation for robust model fusion in federated learning.
\newblock In \emph{Advances in Neural Information Processing Systems}, pages 2351--2363. Curran Associates, Inc., 2020.

\bibitem[Liu et~al.(2023)Liu, Zhang, Qiu, Xie, Zhang, and Yao]{POP}
Sun-Ao Liu, Yiheng Zhang, Zhaofan Qiu, Hongtao Xie, Yongdong Zhang, and Ting Yao.
\newblock Learning orthogonal prototypes for generalized few-shot semantic segmentation.
\newblock In \emph{Proceedings of the IEEE/CVF Conference on Computer Vision and Pattern Recognition}, pages 11319--11328, 2023.

\bibitem[Luo et~al.(2021)Luo, Chen, Hu, Zhang, Liang, and Feng]{nofearofheterogenity}
Mi Luo, Fei Chen, Dapeng Hu, Yifan Zhang, Jian Liang, and Jiashi Feng.
\newblock No fear of heterogeneity: Classifier calibration for federated learning with non-iid data.
\newblock \emph{Advances in Neural Information Processing Systems}, 34:\penalty0 5972--5984, 2021.

\bibitem[Ma et~al.(2022)Ma, Zhu, Lin, Chen, and Qin]{model-training-on-noniid-data}
Xiaodong Ma, Jia Zhu, Zhihao Lin, Shanxuan Chen, and Yangjie Qin.
\newblock A state-of-the-art survey on solving non-iid data in federated learning.
\newblock \emph{Future Generation Computer Systems}, 135:\penalty0 244--258, 2022.

\bibitem[McMahan et~al.(2017)McMahan, Moore, Ramage, Hampson, and y~Arcas]{fedavg}
Brendan McMahan, Eider Moore, Daniel Ramage, Seth Hampson, and Blaise~Aguera y Arcas.
\newblock Communication-efficient learning of deep networks from decentralized data.
\newblock In \emph{Artificial intelligence and statistics}, pages 1273--1282. PMLR, 2017.

\bibitem[Nilsback and Zisserman(2008)]{flowers102}
Maria-Elena Nilsback and Andrew Zisserman.
\newblock Automated flower classification over a large number of classes.
\newblock In \emph{2008 Sixth Indian conference on computer vision, graphics \& image processing}, pages 722--729. IEEE, 2008.

\bibitem[Ranasinghe et~al.(2021)Ranasinghe, Naseer, Hayat, Khan, and Khan]{opl}
Kanchana Ranasinghe, Muzammal Naseer, Munawar Hayat, Salman Khan, and Fahad~Shahbaz Khan.
\newblock Orthogonal projection loss.
\newblock In \emph{Proceedings of the IEEE/CVF international conference on computer vision}, pages 12333--12343, 2021.

\bibitem[Sandler et~al.(2018)Sandler, Howard, Zhu, Zhmoginov, and Chen]{mobilenet}
Mark Sandler, Andrew Howard, Menglong Zhu, Andrey Zhmoginov, and Liang-Chieh Chen.
\newblock Mobilenetv2: Inverted residuals and linear bottlenecks.
\newblock In \emph{Proceedings of the IEEE conference on computer vision and pattern recognition}, pages 4510--4520, 2018.

\bibitem[Shen et~al.(2020)Shen, Zhang, Jia, Zhang, Huang, Zhou, Kuang, Wu, and Wu]{fml}
Tao Shen, Jie Zhang, Xinkang Jia, Fengda Zhang, Gang Huang, Pan Zhou, Kun Kuang, Fei Wu, and Chao Wu.
\newblock Federated mutual learning.
\newblock \emph{arXiv preprint arXiv:2006.16765}, 2020.

\bibitem[Szegedy et~al.(2015)Szegedy, Liu, Jia, Sermanet, Reed, Anguelov, Erhan, Vanhoucke, and Rabinovich]{googlenet}
Christian Szegedy, Wei Liu, Yangqing Jia, Pierre Sermanet, Scott Reed, Dragomir Anguelov, Dumitru Erhan, Vincent Vanhoucke, and Andrew Rabinovich.
\newblock Going deeper with convolutions.
\newblock In \emph{Proceedings of the IEEE conference on computer vision and pattern recognition}, pages 1--9, 2015.

\bibitem[Tan and Le(2019)]{tan2019efficientnet}
Mingxing Tan and Quoc Le.
\newblock Efficientnet: Rethinking model scaling for convolutional neural networks.
\newblock In \emph{International conference on machine learning}, pages 6105--6114. PMLR, 2019.

\bibitem[Tan et~al.(2022{\natexlab{a}})Tan, Long, Liu, Zhou, Lu, Jiang, and Zhang]{tan2022fedproto}
Yue Tan, Guodong Long, Lu Liu, Tianyi Zhou, Qinghua Lu, Jing Jiang, and Chengqi Zhang.
\newblock Fedproto: Federated prototype learning across heterogeneous clients.
\newblock In \emph{Proceedings of the AAAI Conference on Artificial Intelligence}, pages 8432--8440, 2022{\natexlab{a}}.

\bibitem[Tan et~al.(2022{\natexlab{b}})Tan, Long, Ma, Liu, Zhou, and Jiang]{FedPCL}
Yue Tan, Guodong Long, Jie Ma, Lu Liu, Tianyi Zhou, and Jing Jiang.
\newblock Federated learning from pre-trained models: A contrastive learning approach.
\newblock \emph{Advances in neural information processing systems}, 35:\penalty0 19332--19344, 2022{\natexlab{b}}.

\bibitem[Thudumu et~al.(2020)Thudumu, Branch, Jin, and Singh]{comprehensive-survey-of-distance}
Srikanth Thudumu, Philip Branch, Jiong Jin, and Jugdutt Singh.
\newblock A comprehensive survey of anomaly detection techniques for high dimensional big data.
\newblock \emph{Journal of Big Data}, 7:\penalty0 1--30, 2020.

\bibitem[Van~der Maaten and Hinton(2008)]{t-sne}
Laurens Van~der Maaten and Geoffrey Hinton.
\newblock Visualizing data using t-sne.
\newblock \emph{Journal of machine learning research}, 9\penalty0 (11), 2008.

\bibitem[Wang et~al.(2023)Wang, Wang, Zhang, and Fu]{aggregating-vs-privacy}
Lianyu Wang, Meng Wang, Daoqiang Zhang, and Huazhu Fu.
\newblock Model barrier: A compact un-transferable isolation domain for model intellectual property protection.
\newblock In \emph{Proceedings of the IEEE/CVF Conference on Computer Vision and Pattern Recognition}, pages 20475--20484, 2023.

\bibitem[Wang et~al.(2024)Wang, Bian, Zhang, Chen, and Xu]{FedPLVM}
Lei Wang, Jieming Bian, Letian Zhang, Chen Chen, and Jie Xu.
\newblock Taming cross-domain representation variance in federated prototype learning with heterogeneous data domains.
\newblock \emph{arXiv preprint arXiv:2403.09048}, 2024.

\bibitem[Wen et~al.(2022)Wen, Jeon, and Huang]{wen2022federated}
Dingzhu Wen, Ki-Jun Jeon, and Kaibin Huang.
\newblock Federated dropout—a simple approach for enabling federated learning on resource constrained devices.
\newblock \emph{IEEE wireless communications letters}, 11\penalty0 (5):\penalty0 923--927, 2022.

\bibitem[Wu et~al.(2022)Wu, Wu, Lyu, Huang, and Xie]{fedkd}
Chuhan Wu, Fangzhao Wu, Lingjuan Lyu, Yongfeng Huang, and Xing Xie.
\newblock Communication-efficient federated learning via knowledge distillation.
\newblock \emph{Nature communications}, 13\penalty0 (1):\penalty0 2032, 2022.

\bibitem[Xu et~al.(2023)Xu, Tong, and Huang]{FedPAC}
Jian Xu, Xinyi Tong, and Shao-Lun Huang.
\newblock Personalized federated learning with feature alignment and classifier collaboration.
\newblock \emph{arXiv preprint arXiv:2306.11867}, 2023.

\bibitem[Ye et~al.(2023)Ye, Fang, Du, Yuen, and Tao]{data_and_model_heterogenous}
Mang Ye, Xiuwen Fang, Bo Du, Pong~C Yuen, and Dacheng Tao.
\newblock Heterogeneous federated learning: State-of-the-art and research challenges.
\newblock \emph{ACM Computing Surveys}, 56\penalty0 (3):\penalty0 1--44, 2023.

\bibitem[Yi et~al.(2023)Yi, Wang, Liu, Shi, and Yu]{yi2023fedgh}
Liping Yi, Gang Wang, Xiaoguang Liu, Zhuan Shi, and Han Yu.
\newblock Fedgh: Heterogeneous federated learning with generalized global header.
\newblock In \emph{Proceedings of the 31st ACM International Conference on Multimedia}, pages 8686--8696, 2023.

\bibitem[Zhang et~al.(2023)Zhang, Guo, Guo, Zeng, Zhou, and Zomaya]{public_dataset_is_difficult_to_obtain}
Jie Zhang, Song Guo, Jingcai Guo, Deze Zeng, Jingren Zhou, and Albert~Y Zomaya.
\newblock Towards data-independent knowledge transfer in model-heterogeneous federated learning.
\newblock \emph{IEEE Transactions on Computers}, 72\penalty0 (10):\penalty0 2888--2901, 2023.

\bibitem[Zhang et~al.(2024)Zhang, Liu, Hua, and Cao]{fedtgp}
Jianqing Zhang, Yang Liu, Yang Hua, and Jian Cao.
\newblock Fedtgp: Trainable global prototypes with adaptive-margin-enhanced contrastive learning for data and model heterogeneity in federated learning.
\newblock In \emph{Proceedings of the AAAI Conference on Artificial Intelligence}, pages 16768--16776, 2024.

\bibitem[Zhu et~al.(2021)Zhu, Hong, and Zhou]{FedGen}
Zhuangdi Zhu, Junyuan Hong, and Jiayu Zhou.
\newblock Data-free knowledge distillation for heterogeneous federated learning.
\newblock In \emph{International conference on machine learning}, pages 12878--12889. PMLR, 2021.

\end{thebibliography}
